\documentclass{article}

\usepackage{arxiv}

\usepackage[utf8]{inputenc} 
\usepackage[T1]{fontenc}    
\usepackage{hyperref}       
\usepackage{url}            
\usepackage{booktabs}       
\usepackage{amsfonts}       
\usepackage{nicefrac}       
\usepackage{microtype}      
\usepackage{lipsum}
\usepackage{graphicx}
\usepackage{orcidlink}
\usepackage{xcolor} 
\usepackage[normalem]{ulem}
\usepackage{color,soul}
\usepackage[table]{xcolor}
\usepackage{wrapfig}
\usepackage{multirow} 
\usepackage{tikz}
\usetikzlibrary{matrix}
\usepackage{pgfplots}
\pgfplotsset{compat=1.18}
\usepackage{amsmath}
\usepackage{subcaption}
\usepackage{cleveref}
\usepackage{svg}
\graphicspath{ {./images/} }

\title{ScratchSim: A Procedural Synthetic Data Pipeline for Surface Scratch Detection}

\author{
Paul Julius Kühn\textsuperscript{*} \\
Fraunhofer IGD, Darmstadt, Germany \And
Saptarshi Neil Sinha\textsuperscript{*} \\
Fraunhofer IGD, Darmstadt, Germany \And
Tiago Kleist \\
Fraunhofer IGD, Darmstadt Germany \And
Richard Hoffmann \\
Fraunhofer IGD,  Darmstadt Germany  \And
Arjan Kuijper \\
Fraunhofer IGD \& TU Darmstadt, Darmstadt, Germany  \And
Michael Weinmann \\
Delft University of Technology, Delft, Netherlands
}

\begin{document}
\maketitle
\renewcommand{\thefootnote}{\fnsymbol{footnote}}
\footnotetext[1]{Equal contribution}

\begin{abstract}
While automated defect detection such as the detection of surface scratched is an important aspect in industrial quality control, the scarcity of annotated defect data make this task challenging. This paper presents a procedural
rendering pipeline that generates large-scale annotated synthetic training data
using BlenderProc, with configurable material appearance, camera modes, and
domain randomization, producing automatic COCO-format annotations. To show the potential of our approach, we evaluate
four training strategies, namely synthetic-only, real-only, mixed, and
fine-tuning from synthetic weights, across two objects with different material
properties and three lightweight edge-deployable detectors, YOLOX, YOLO26,
and LW-DETR. Our evaluation show that fine-tuning from synthetic weights consistently
outperforms real-only training, and that mixed training effectively recovers
performance under scarce real-data conditions, with findings validated across
both convolutional and transformer-based architectures. The proposed approach
enables scalable defect detection without the burden of large real annotated
datasets, making it practical for on-device industrial inspection. The pipeline scripts for generating synthetic scratches, 3D model, and both the synthetic and real annotated scratch
datasets for a glossy toy Ferrari car are publicly available at
\url{https://github.com/saptarshineil/ScratchSim}.
\end{abstract}
\section{Introduction}
\label{sec:introduction}
Industrial quality control in large-scale manufacturing relies on reliable and fast 
defect detection. Effective defect detection raises product standards in terms of 
service life, performance, and safety, while its application in early production 
stages can reduce risk, waste, and rework, contributing positively to efficiency and 
sustainability~\cite{qiao2025metal}. Although manual visual inspection remains 
prevalent, the need for automated solutions is well established. Human inspection, 
while flexible and intuitive, is labor-intensive, difficult to scale, and inherently 
inconsistent, with fatigue adversely affecting accuracy~\cite{NEWMAN1995231}. As 
modern manufacturing continues to increase production speed and throughput, manual 
inspection becomes more and more infeasible~\cite{zhou2019automobile}.

\begin{figure}[htb!]
  \centering
  \includegraphics[width=1.0\textwidth]{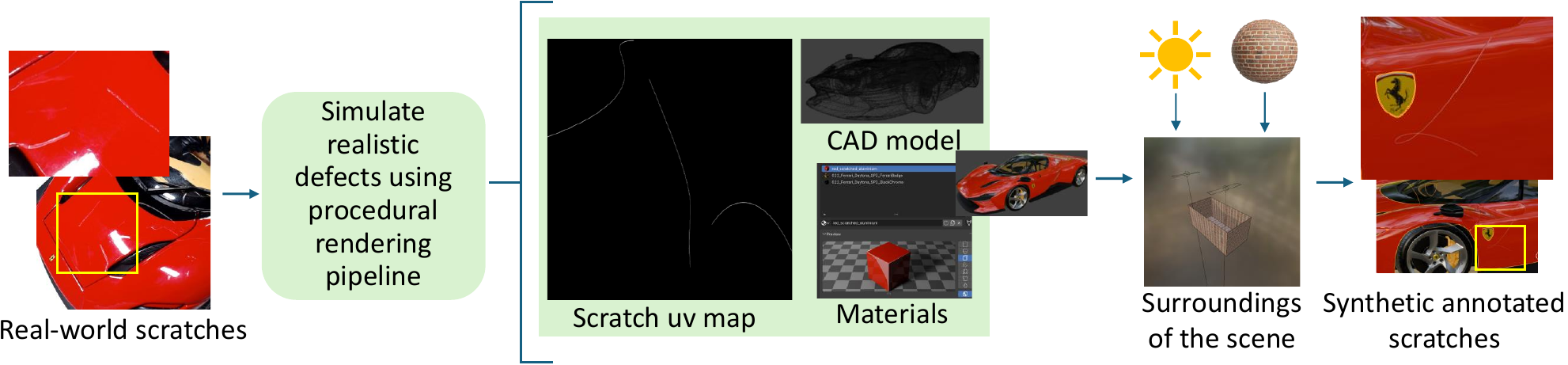}
  \caption{Overview of the proposed procedural rendering pipeline utilizing Blenderproc~\cite{denninger2023blenderproc2} for synthetic scratch data generation. A CAD model is combined with procedurally generated scratch UV maps, PBR materials, and randomized scene surroundings to produce photorealistic images with automatic pixel-level annotations, guided by the appearance of real-world scratches.}
  \label{fig:pipeline_synthetic_defect_generation}
\end{figure}

Rapid advances in deep learning have shown great potential for 
automated defect detection, offering high accuracy and real-time performance. However, 
these capabilities depend on large amounts of annotated training data. In practice, 
defect samples are rare, with only a small fraction of production output depicting 
defects~\cite{app11167657}. This limited data availability makes models prone to 
overfitting, reduced accuracy, and poor generalization. Synthetic training data 
is a promising way to address this problem.
Procedural rendering pipelines~\cite{denninger2019blenderproc,denninger2023blenderproc2} allow the generation of large quantities of fully annotated 
images that can replace or supplement real data. This approach gives full control over data quantity, quality, and content, 
but synthetic images do not always represent real-world conditions accurately 
enough. Synthetic data has been successfully applied across various industrial 
tasks using procedural rendering frameworks such as 
BlenderProc~\cite{denninger2023blenderproc2} and NVIDIA Isaac 
Sim~\cite{gao2026nvidiaisaacsimenabling}, and combining synthetic with real data has shown 
promise in bridging this gap~\cite{mumuni2024survey, app11167657}. Increasing 
realism and applying domain randomization are two common strategies to further 
improve synthetic-to-real transfer, yet how to best configure and use synthetic 
data remains an open question~\cite{jain2022synthetic}. To address this in terms of generating realistic surface scratch data and identifying plausible training configurations, this paper presents a procedural pipeline for generating annotated synthetic data for surface scratch detection, applied to two objects with distinct material properties: a glossy toy Ferrari car and a matte powder-coated industrial grip.
We generated multiple synthetic datasets, two for a glossy toy Ferrari car and 
four for a matte industrial grip, utilizing varying pipeline configurations, 
and used them to benchmark four training strategies across three lightweight 
edge-deployable detectors, with all models evaluated on real images to 
assess synthetic-to-real transfer. In summary, the key contributions of this work are:
\begin{itemize}
    \item A procedural rendering pipeline (see Figure~\ref{fig:pipeline_synthetic_defect_generation}) for generating annotated synthetic
    training data for surface scratch detection, supporting configurable
    material appearance, camera modes, and domain randomization, with
    annotations automatically produced in COCO format~\cite{coco}.
    \item We present a comparative evaluation of four training strategies (synthetic-only, 
real-only, mixed synthetic and real, and  fine-tuning a synthetically pretrained model on real data) tested 
on three lightweight edge-deployable detectors (YOLOX~\cite{yolox2021}, 
YOLO26~\cite{jocher2026ultralyticsyolo26unifiedrealtime}, and 
LW-DETR~\cite{chen2024lw}) and two objects with distinct material properties. 
    \item Publicly released scratch datasets for a glossy toy Ferrari car, comprising both real-world manually annotated images and synthetically generated samples, to support future research. We do not release the industrial grip datasets due to intellectual property 
restrictions.
\end{itemize}

\section{Related works}
\label{sec:related_works}
Industrial defect detection has shifted from traditional computer vision
methods towards deep learning, as classical approaches are sensitive to noise,
lighting, and complex backgrounds~\cite{11006646}. CNNs have improved on these
limitations through superior feature extraction and adaptability~\cite{11006646,
qiao2025metal}, yet supervised methods still require large, well-labelled
datasets that are costly and scarce in industrial settings~\cite{a16020095}.
Data augmentation techniques such as flipping, rotation, and scaling are
widely used to combat overfitting and expand small datasets~\cite{mumuni2024survey,
app11167657}. While effective, augmentation introduces no genuinely new
information, leaving performance still dependent on the quantity and quality
of the original data~\cite{jain2022synthetic}. Synthetic data generation has
emerged as a more complete solution to data scarcity, with early work by
Shotton et al.~\cite{shotton} demonstrating that large synthetic datasets can
enable real-time pose recognition invariant to appearance variation, motivating
broader adoption across vision tasks. Methods broadly span GAN-based image synthesis, 
diffusion models, and 3D rendering pipelines~\cite{arlovic2024synthetic}. 
GAN-based approaches can produce diverse images but struggle with
precise annotation control and require large training sets
themselves~\cite{DEMELO2022174,MONNET2024767}. Diffusion models such as Stable Diffusion~\cite{rombach2022stablediffusion} and DALL-E 3~\cite{betker2023dalle3} have shown impressive image quality, yet similarly offer limited control over precise object-level annotations 
and geometric scene properties. Recent work by Kühn et 
al.~\cite{kuhn2026synsurendtoendgenerativepipeline} explores an end-to-end 
diffusion-based pipeline for industrial surface defect synthesis, finding 
that synthetic-only training does not replace real data but can yield modest 
gains when combined with it. While both generative approaches show promise 
for data augmentation, a thorough exploration of generative AI for 
controllable industrial defect synthesis lies beyond the scope of this work. Instead, 3D rendering pipelines offer full scene control, automatic annotation, and 
independence from real data. Prior work by Weinmann et
al.~\cite{weinmann_synthetic} demonstrated that synthesizing training data from
measured material and illumination characteristics enables material
classification under complex real-world conditions, highlighting the value of
appearance-faithful synthetic rendering. Frameworks such as 
BlenderProc~\cite{denninger2019blenderproc,denninger2023blenderproc2}, as 
demonstrated for 6D pose estimation in~\cite{sinha20256dstrawberryposeestimation}, 
and GPU-accelerated simulators such as NVIDIA Isaac 
Sim~\cite{gao2026nvidiaisaacsimenabling} simplify the creation of annotated synthetic 
datasets at scale. Domain randomization
further reduces the domain gap by randomizing non-essential scene properties,
improving generalization without the effort of photorealistic
rendering~\cite{tobin2017DR,8575297,carla,cabon2020vkitti2}. Several strategies exist for 
incorporating synthetic data into training. Models
trained solely on synthetic data often underperform due to the domain gap,
motivating combined approaches that use small amounts of real data through
hybrid datasets or fine-tuning~\cite{seib2020mixingrealsyntheticdata}.
Kim et al.~\cite{KIM2023104771} showed that hybrid training can achieve
near-maximum performance with up to an 80\% reduction in real data. 
Regarding detection algorithms, single-stage detectors such as the YOLO
series~\cite{yolo,yolox2021,jocher2026ultralyticsyolo26unifiedrealtime} and
Single Shot MultiBox Detector (SSD)~\cite{Liu2016Ssd} offer fast inference with low computational requirements,
making them well suited for real-time defect detection on edge
devices~\cite{machines2023yolo1-8}. More recently, transformer-based
single-stage detectors such as LW-DETR~\cite{chen2024lw} have shown
competitive real-time performance while offering a viable alternative to
CNN-based approaches. Two-stage methods such as the R-CNN
series~\cite{rcnn,fast_rcnn,faster_rcnn} prioritise accuracy at the cost of
higher latency, making them less practical for on-device deployment. Given the practical requirement for real-time inference on edge devices, 
single-stage lightweight detectors are the natural choice for this work, 
and hence we benchmark YOLOX~\cite{yolox2021}, 
YOLO26~\cite{jocher2026ultralyticsyolo26unifiedrealtime}, and 
LW-DETR~\cite{chen2024lw}, spanning both CNN- and transformer-based 
architectures. For procedural rendering, we adopt BlenderProc~\cite{denninger2023blenderproc2}, 
an open-source framework that addresses the data scarcity challenge outlined 
above through built-in domain randomization.
\section{Methodology}
\label{sec:methodology}
In this section, we present our approach for the generation of annotated synthetic high-quality training data for scratch detection.
\subsection{Synthetic Data Generation Pipeline}
\label{subsec:pipeline}
\begin{figure}[htbp]
    \centering
    \includegraphics[width=\textwidth]{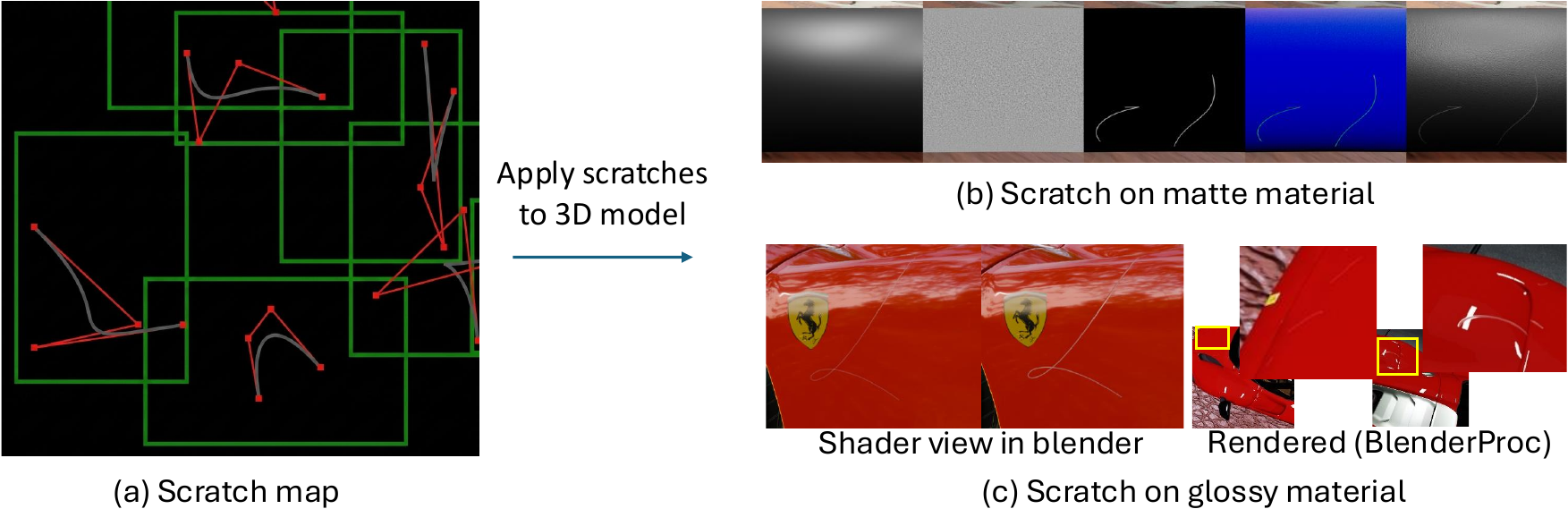}
    \caption{Scratch generation and application. (a) Generated scratch map with annotations. 
(b) Stages of scratch application on a matte material (left to right): blank object, surface bumps,
applied scratch mask, resulting normal map, and final result. (c) Effect of scratch 
intensity on a glossy material (left to right): strength 0.1 and strength 1.0.}
    \label{fig:scratch_generation}
\end{figure}

We built our synthetic data generation pipeline on BlenderProc~\cite{denninger2023blenderproc2}, a modular procedural rendering framework based on Blender and its Python API. It was chosen for its open-source availability, strong documentation, active community support, and suitability for scientific use. BlenderProc~\cite{denninger2023blenderproc2} provides key components for scalable synthetic data generation while retaining the flexibility to implement application-specific functionality through the underlying Blender API. An overview of our pipeline for generating synthetic defect data for scratches is shown in \cref{fig:pipeline_synthetic_defect_generation}. The following subsections describe details of our synthetic defect data generation pipeline.
\noindent\textbf{Geometry preparation}
\label{subsubsec:object_preparation}
The object of interest is represented by a CAD model prepared for realistic rendering. Geometric refinements, such as bevels and subdivision surfaces, are applied where needed to enhance visual realism. The model is scaled to a 1:1 ratio using the known physical dimensions of the real object. A UV map is then generated to support the projection of both the base material and the procedurally generated scratch maps. If the model contains regions that are not intended to appear in the final renders, these are isolated into separate UV islands by placing seams accordingly.
 \begin{figure}[htb!]
  \centering
  \includegraphics[width=\textwidth]{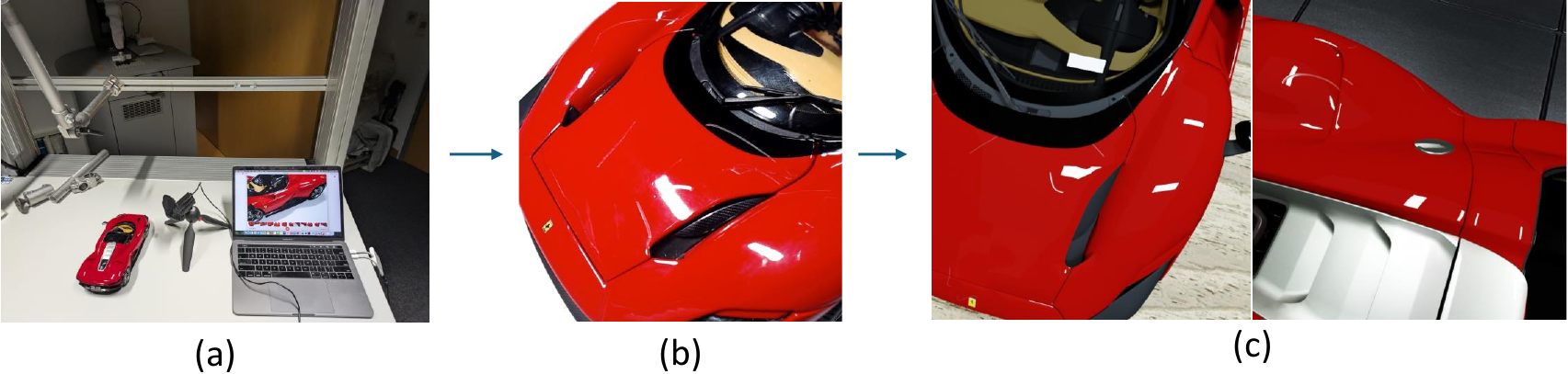}
  \caption{Application of the material pipeline to a toy Ferrari~\cite{car_model1,car_model2}, used as the car object in the synthetic scratch datasets. (a) Real data capture setup, (b) real image of the toy Ferrari, and (c) simulated material (multiple rendered views) applied in BlenderProc~\cite{denninger2023blenderproc2} with adjusted reflectance and surface roughness parameters to match the glossy automotive finish.}
  \label{fig:material_capture_car}
\end{figure}
\noindent\textbf{Material construction}
\label{subsubsec:material_construction}
The material is constructed procedurally and comprises three layers: a surface appearance model, procedurally generated scratches, and their interaction.
\noindent\textit{Surface appearance modeling.}
The surface appearance is modeled procedurally to approximate the real 
object's material characteristics. As a first object, we considered an 
industrial grip with a matte, powder-coated surface. Since a powder-coated surface exhibits surface irregularities at multiple scales, with fine granular roughness from the coating particles, two noise textures varying in 
scale and level of detail are mixed to capture both frequency components 
and achieve a visually representative result.
The combined noise texture is used to generate a height map, from which a 
normal map is derived and applied to the object via UV mapping (see Figure~\ref{fig:scratch_generation}(b)).
For glossy automotive surfaces, the surface appearance is governed less by 
microscale geometry and more by the optical properties of the paint system 
itself, which typically consists of a pigmented base coat and a transparent 
clear coat contributing high specularity. This multi-layer characteristic is 
approximated by adjusting the material parameters towards higher specularity, 
lower surface roughness, and increased reflectance, without requiring noise 
blending. This allows the pipeline to reproduce the visually distinct 
appearance of an automotive finish, as shown in 
\cref{fig:material_capture_car}.
\noindent\textit{Scratch generation.}
Scratches are synthesized procedurally on an $8192 \times 8192$ black canvas,
with $N \sim \mathcal{U}[i, j]$ scratches per mask, where $i$ and $j$ depend on
the object's surface size and the desired scratch density, ensuring variety
across training samples.
Each scratch is defined by a bounding box of random size
$s_\mathrm{box} \in \left[\frac{s_\mathrm{canvas}}{k}, \frac{s_\mathrm{canvas}}{l}\right]$,
placed at a random position within the canvas, where the scaling factors $k, l$
are chosen to reflect the typical physical size of real scratches observed on the
objects. For the toy car, scratch sizes are instead drawn from a weighted
distribution over predetermined values, mirroring the distribution of its real
scratches. Within each box, four randomly sampled points serve as start point,
end point, and two control points of a cubic B\'ezier curve, approximating the
irregular, curved geometry of real scratches. After rendering all curves, a
Gaussian blur softens the edges and suppresses aliasing, and the image is
downscaled by a factor of two and normalized, reducing the memory footprint while
yielding smooth, consistent intensity values. Figure~\ref{fig:scratch_generation}(a)
visualizes a resulting scratch mask with annotations to aid understanding of the
following workflow.
\begin{wrapfigure}{r}{0.28\linewidth}
 \vspace{10pt}
    \centering
    \includegraphics[width=\linewidth]{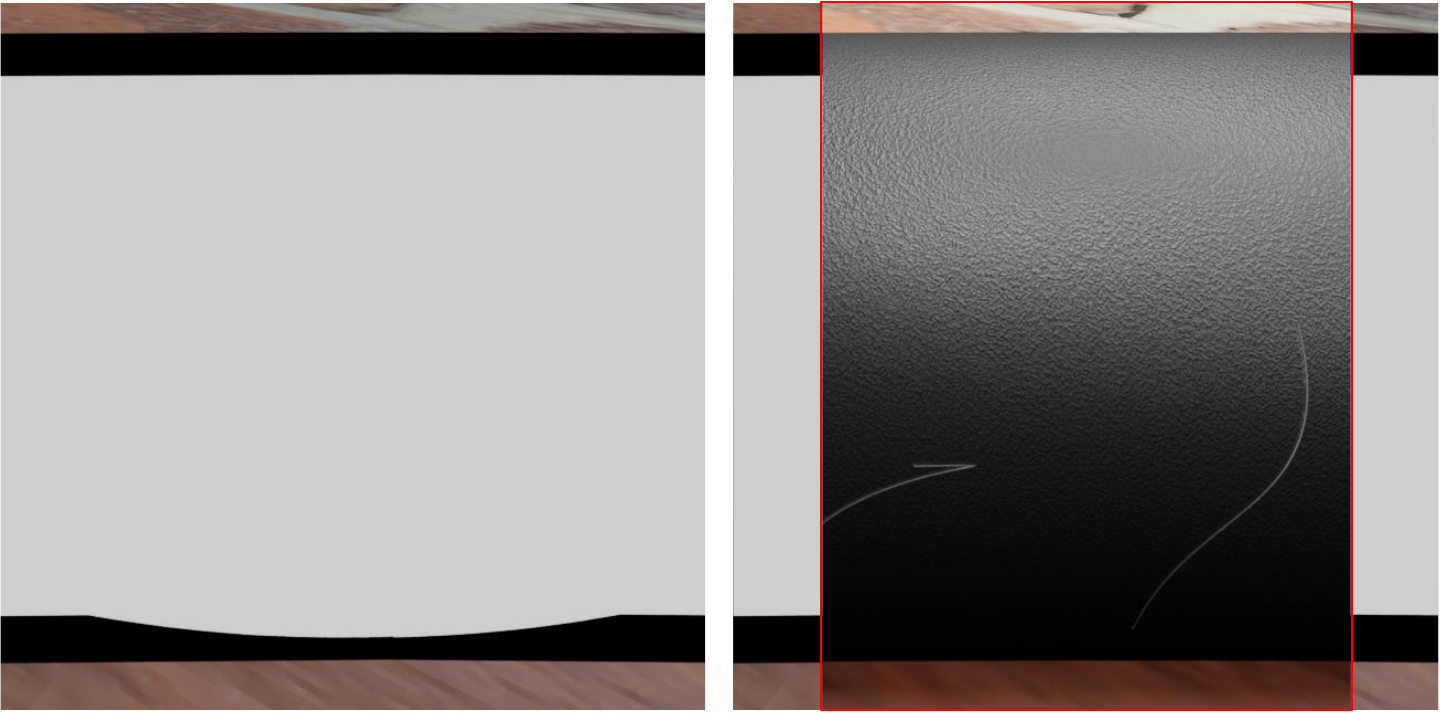}
    \caption{Allowed (white) and disallowed (black) viewing angles for scratch visibility (left). Overlaid original rendering on mask (right) to show that scratches generated as desired.}
    \label{fig:angle_restriction}
\end{wrapfigure}
\noindent\textit{Viewing-Angle restriction.}
While rendering scratches on round objects, the viewing angle near the 
edges becomes too shallow to reveal the scratch geometry, resulting in 
scratches that are invisible in the rendered image yet still annotated, 
introducing false positives into the training data. To address this, 
scratches at excessively shallow viewing angles are suppressed during 
rendering. This is implemented in the Blender material nodes by computing 
the dot product of the unit surface normal $\vec{n}$ and the unit 
camera viewing direction $\vec{v}$, such that $|\vec{v} \cdot \vec{n}| > t$,
where $t \in [0, 1]$ is a user-defined threshold controlling 
the maximum permitted viewing angle relative to the surface normal.
The resulting restriction is visualized in Figure~\ref{fig:angle_restriction}, where white regions indicate 
areas in which scratches are rendered and black regions indicate suppressed areas.
%

%
\noindent\textit{Scratch application}
The proposed scratch generation approach is object-agnostic and can be 
applied to any 3D object, provided an appropriate material approach is 
selected based on the surface type. We applied our approach on two materials as presented in Figure~\ref{fig:scratch_generation}(b) and~(c). For coated or painted matte surfaces, shown in 
Figure~\ref{fig:scratch_generation}(b), the scratch mask value 
$v_{\text{scratch}}$ and a randomised depth factor $d_{\text{scratch}} = 
v_{\text{scratch}} \cdot r$, with $r \in [0, 1]$, modulate multiple 
material properties simultaneously. For roughness, $v_{\text{scratch}}$ is 
added to the base roughness (clamped to $[0,1]$), simulating surface 
roughening independent of depth. For colour, $d_{\text{scratch}}$ mixes 
the surface colour toward a metallic grey, representing exposed substrate. 
The metallic component is increased by $2 \cdot d_{\text{scratch}}$, 
clamped above the base value, to simulate bare metal exposure at greater 
depths. Surface normals are attenuated by $v_{\text{scratch}}$ to simulate 
removal of the coating texture, while scratch normals are blended in 
proportional to $d_{\text{scratch}}$. Finally, a small negative 
displacement proportional to $d_{\text{scratch}}$ is applied for geometric 
realism.

For smooth glossy or reflective surfaces, shown in 
Figure~\ref{fig:scratch_generation}(c), a simpler approach is used, where 
the scratch mask is applied as a normal map perturbation with a controllable 
strength parameter, allowing the depth and visibility of scratches to be 
varied. In addition, colour modifications are performed in HSV space for 
better control over scratch appearance.
\noindent\textbf{Camera Positioning}
\label{subsubsec:camera_positioning}
To ensure comprehensive scene coverage and capture the full range of 
lighting and material appearance changes across the object surface, 
the camera viewpoint is varied for each frame. For each frame, a target point is sampled uniformly on the object's visible surface by selecting a polygon with probability proportional to its area, then sampling a random point on the selected quad via bilinear interpolation with two uniform parameters $l_1, l_2 \in [0, 1)$.
Two camera modes are implemented. In the close-range mode (\texttt{randomcam}), the camera is placed on a spherical shell centred on the target point with radius $r \in [2, 5]$~cm, altitude $\beta \in [30°, 90°]$, and azimuth constrained within $\pm 20°$ of the surface normal. In the tripod mode, the camera height is sampled in $[12, 16]$~cm, with horizontal position sampled on a disk of radius 7.5~cm centred on the target point.
The tripod mode was used exclusively for the glossy toy car object, reflecting a more controlled, fixed-distance acquisition setup. For the matte industrial object (grip), both modes were employed to provide a greater variety of viewpoints and scales. Each candidate pose undergoes validity checks: (1)~camera lies within room boundaries, (2)~no obstacle within 1~cm of the lens, (3)~neither camera position nor near-frustum vertices lie inside the object mesh (verified via dot-product sign test against closest-surface normals), and (4)~excluded surface regions are not visible (verified via frustum containment and ray casting). Invalid poses are resampled.
\noindent\textbf{Environment and Lighting}
\label{subsubsec:scene_composition}
The design of the object's surroundings was heavily influenced by the implementation 
of~\cite{hodan2020bopchallenge20206d}, which generates photorealistic datasets for 
6D object-pose estimation using BlenderProc~\cite{denninger2023blenderproc2}. An empty room is constructed from five 
planes (four walls and a floor), all sharing a single PBR material chosen at random 
for each scene from the texture library provided by~\cite{ambientCG}. Two fixed overhead light sources, positioned on opposite ends of the room and slightly offset in 
the same direction we used to illuminate the scene. Each light can be controlled independently 
in colour and intensity, producing a wide variety of lighting conditions that also 
depend on the object's position within the room. This configuration simplifies ensuring 
sufficient visibility by constraining light intensities accordingly, which proved 
necessary as heavily randomised lighting caused details to be obscured by over- or 
underexposure in initial testing. This domain-randomisation-centred approach, with varying materials and lighting across scenes, is intended to shift the model's focus towards the object of interest and improve robustness to environmental changes, while avoiding the need to model a specific predetermined location. The resulting environment and lighting setup can also be seen as part of the full 
pipeline in Figure~\ref{fig:pipeline_synthetic_defect_generation}.
\noindent\textbf{Automatic annotation}
\label{subsubsec:automatic_annotation}
Annotations are generated without manual intervention by utilizing an Arbitrary Output Variable (AOV) node that exports the applied scratch mask as rendered from the camera's perspective. The AOV output is binarised (thresholding near-black pixels), and connected components are labelled to identify individual scratch instances. The resulting per-pixel segmentation is passed to a COCO-format~\cite{coco} annotation writer, which automatically computes bounding boxes for each labelled region, producing detection-ready ground truth alongside each rendered image.
\noindent\textbf{Pipeline execution}
\label{subsubsec:pipeline_execution}
The pipeline follows a nested loop controlled by two parameters, the number of 
scenes $S$ and the number of frames per scene $F$, producing $S \times F$ images 
in total. For each scene, the scratch mask, material properties, object pose, 
background, and lighting are randomised. Within a scene, only the camera pose 
changes between frames. This keeps scene generation efficient while still producing 
diverse viewpoints. All images are rendered at $640 \times 640$ pixels using 
Blender's Cycles engine with PBR shading.

\section{Comparison using edge-deployable baseline detectors}
\label{sec:detector_models}
To validate the suitability of our generated synthetic training data for defect detection, three lightweight real-time object detectors are used across the two object's datasets. For the industrial grip, we use YOLOX-S~\cite{yolox2021}, an open-source convolutional detector chosen specifically because its permissive license allows commercial deployment. We evaluated two models for the toy Ferrari car datasets, YOLO26-n~\cite{jocher2026ultralyticsyolo26unifiedrealtime}, a small convolutional detector from the latest YOLO generation designed for fast and accurate detection, and LW-DETR-Tiny~\cite{chen2024lw}, a compact transformer-based detector that achieves real-time performance while offering a fundamentally different architecture, 
allowing findings to be assessed across both model families.
\section{Generated datasets}
\label{sec:datasets}
Datasets were created for two objects, a matte industrial grip and a glossy Ferrari 
toy car (see Figure~\ref{fig:scratch_generation}), to evaluate the impact of synthetic data on model performance. For each 
object, both synthetic and real datasets were produced. All synthetic datasets 
contain 10\,000 images each, split into 7\,000 training, 2\,000 validation, and 
1\,000 test images following a 70/20/10 ratio. Images are rendered at 
$640 \times 640$ pixels with five frames per scene. All real images were captured 
using a Logitech HD Pro C920 webcam, cropped to $640 \times 640$ pixels, and 
manually annotated in COCO format~\cite{coco}. It includes both clean and freshly scratched examples to test detection performance and measure false-positive rates.
\noindent\textbf{Ferrari toy car datasets}
Two synthetic datasets were generated for the Ferrari toy car using 
the tripod camera mode and the glossy material configuration described in 
\cref{subsubsec:material_construction}, featuring either a randomized or white static background (modeled) similar to the capture setup for automatic visual inspection. The real dataset contains 116 images 
captured in front of static background under different lighting conditions and 
camera angles, split into 82 training, 18 validation, and 16 test images.
\noindent\textbf{Industrial grip datasets}
To increase scene variation and study the effect of color and camera 
randomization on model generalization, four synthetic base datasets were 
created for the industrial grip by combining two color modes, 
\texttt{tricolour} and \texttt{randomcolour}, with two camera modes, 
\texttt{randomcam} and \texttt{tripod}. The \texttt{tricolour} mode 
selects one of the three real color variants of the grip for each scene, 
closely matching the object's actual appearance. The other mode \texttt{randomcolour} instead randomizes the surface color and 
roughness for each scene, increasing visual diversity and making the model 
less dependent on a specific color. Two real datasets were captured for 
the grip. The first contains 120 images taken in front of five different 
backgrounds under varied lighting and camera angles, used for training. 
The second contains 120 images taken against a static background under 
controlled lighting conditions, used exclusively for evaluation.
\section{Implementation details}
All training and evaluation was performed in a Docker container running within a SLURM job on a GPU cluster containing NVIDIA A100 SXM4 40\,GB GPU's (4 GPUs for LW-DETR and 1 GPU for the YOLO series of detectors). 
The YOLOX-S model~\cite{yolox2021} was trained on the industrial grip datasets,
using a batch size of 64 for 125 epochs with validation every 10 epochs for
synthetic and mixed runs, and a batch size of 4 for 50 epochs with validation
every epoch for fine-tuning, with the default YOLOX learning rate scheduler and
augmentation pipeline. The epoch counts were determined from preliminary experiments
showing AP stagnation around these points.
For the toy Ferrari dataset, YOLO26~\cite{jocher2026ultralyticsyolo26unifiedrealtime} 
and LW-DETR~\cite{chen2024lw} were each trained with batch sizes scaled 
proportionally to the dataset size to maintain stable gradient estimates: 
scarce real-data splits used $\text{bs}=2$ (10\%), $\text{bs}=4$ (25\%), 
and $\text{bs}=8$ (50\%), while full-dataset and mixed-data runs used 
$\text{bs}=16$ (real 100\% and fine-tuning) or $\text{bs}=32$ 
(synthetic-only and all infused configurations). YOLO26 was trained for 
150--300 epochs depending on the regime (150 for real-only and fine-tuning, 
200 for synthetic baselines, 300 for infused mixes), whereas LW-DETR used 
a fixed 90-epoch schedule throughout. To account for training variability, 
both models were evaluated across three independent runs, with results 
reported as mean~$\pm$~standard deviation.
During synthetic dataset generation, the visibility threshold $t$ was 
determined empirically through visual inspection and set to $0.4$ for 
the industrial grip and $0.55$ for the toy car.

\section{Evaluation}
This section presents the quantitative and qualitative results of all conducted experiments across both objects (see Figure~\ref{fig:scratch_generation}).
\begin{table}[htbp]
\centering
\caption{Detection results on the toy Ferrari datasets for YOLO26~\cite{jocher2026ultralyticsyolo26unifiedrealtime}
and LW-DETR~\cite{chen2024lw}.
\colorbox {yellow!40}{Yellow} = best result,
\colorbox{gray!50}{Gray} = second best (comparing mean values).
\textbf{WB} =  synthetic images rendered with a white background similar to real image capture setup (default: random background).}
\label{tab:combined_results}
\resizebox{\textwidth}{!}{%
\begin{tabular}{lc!{\color{black!60}\vrule width 1.0pt}c!{\color{black!60}\vrule width 1.0pt}c!{\color{black!60}\vrule width 1.0pt}c!{\color{black!60}\vrule width 1.0pt}c!{\color{black!60}\vrule width 1.0pt}c}
\toprule
& \multicolumn{4}{c}{\textbf{YOLO26}~\cite{jocher2026ultralyticsyolo26unifiedrealtime}}
& \multicolumn{2}{c}{\textbf{LW-DETR}~\cite{chen2024lw}} \\
\cmidrule(lr){2-5}\cmidrule(lr){6-7}
\textbf{Training Regimes}
  & \textbf{mAP50} & \textbf{mAP50-95} & \textbf{Precision}$^\dagger$ & \textbf{Recall}$^\dagger$
  & \textbf{mAP50} & \textbf{mAP50-95} \\
\midrule
\multicolumn{7}{l}{\textit{Baselines}} \\[2pt]
\quad Real (100\%)
  & 0.6100 ± 0.0255 & 0.4024 ± 0.0040 & 0.7024 ± 0.0623 & 0.5489 ± 0.0042
  & 0.5350 ± 0.0303 & 0.3657 ± 0.0080 \\
\quad Synthetic
  & 0.4513 ± 0.0282 & 0.3000 ± 0.0199 & 0.4828 ± 0.0650 & 0.4683 ± 0.0297
  & 0.3817 ± 0.0129 & 0.2430 ± 0.0108 \\
\quad Synthetic (WB)
  & 0.4067 ± 0.0244 & 0.2590 ± 0.0237 & 0.4811 ± 0.0653 & 0.3824 ± 0.0674
  & 0.3260 ± 0.0111 & 0.2023 ± 0.0091 \\
\midrule
\multicolumn{7}{l}{\textit{Real Only — Scarce}} \\[2pt]
\quad Real (10\%)
  & 0.0561 ± 0.0076 & 0.0207 ± 0.0030 & 0.0053 ± 0.0008 & 0.3775 ± 0.0595
  & 0.2380 ± 0.0234 & 0.1507 ± 0.0275 \\
\quad Real (25\%)
  & 0.0428 ± 0.0053 & 0.0162 ± 0.0046 & 0.1198 ± 0.1012 & 0.1274 ± 0.0810
  & 0.3553 ± 0.0272 & 0.2433 ± 0.0101 \\
\quad Real (50\%)
  & 0.3229 ± 0.0364 & 0.2119 ± 0.0264 & 0.3771 ± 0.0240 & 0.3841 ± 0.0381
  & 0.4070 ± 0.0397 & 0.2617 ± 0.0466 \\
\midrule
\multicolumn{7}{l}{\textit{Synth + Real Mixed (Infused)}} \\[2pt]
\quad Synth + Real (10\%)
  & 0.5065 ± 0.0385 & 0.3431 ± 0.0293 & 0.6221 ± 0.1220 & 0.4781 ± 0.0242
  & 0.3873 ± 0.0121 & 0.2470 ± 0.0090 \\
\quad Synth (WB) + Real (10\%)
  & 0.4319 ± 0.0115 & 0.2767 ± 0.0068 & 0.5526 ± 0.0535 & 0.4314 ± 0.0516
  & 0.3867 ± 0.0187 & 0.2387 ± 0.0060 \\
\quad Synth + Real (25\%)
  & 0.5479 ± 0.0496 & 0.3760 ± 0.0309 & 0.6232 ± 0.0287 & 0.5424 ± 0.0239
  & 0.5120 ± 0.0113 & 0.3407 ± 0.0095 \\
\quad Synth (WB) + Real (25\%)
  & 0.5979 ± 0.0259 & 0.4327 ± 0.0169 & 0.5793 ± 0.0250 & 0.6176 ± 0.0736
  & 0.5113 ± 0.0327 & 0.3320 ± 0.0161 \\
\quad Synth + Real (50\%)
  & 0.6345 ± 0.0197 & 0.4291 ± 0.0254 & 0.6914 ± 0.0475 & 0.5686 ± 0.0306
  & 0.6067 ± 0.0154 & 0.4180 ± 0.0061 \\
\quad Synth (WB) + Real (50\%)
  & 0.6722 ± 0.0023 & 0.4951 ± 0.0096 & 0.7384 ± 0.0739 & 0.6192 ± 0.0449
  & 0.6357 ± 0.0246 & 0.4220 ± 0.0078 \\
\quad Synth + Real (100\%)
  & 0.6913 ± 0.0327 & 0.4628 ± 0.0219 & \cellcolor {yellow!40}0.8280 ± 0.0218 & 0.6197 ± 0.0371
  & 0.6633 ± 0.0159 & 0.4577 ± 0.0091 \\
\quad Synth (WB) + Real (100\%)
  & \cellcolor{gray!50}0.7064 ± 0.0083 & \cellcolor{gray!50}0.5038 ± 0.0085 & \cellcolor{gray!50}0.8085 ± 0.0743 & \cellcolor{gray!50}0.6405 ± 0.0174
  & 0.6830 ± 0.0221 & 0.4593 ± 0.0211 \\
\midrule
\multicolumn{7}{l}{\textit{Fine-tuning (Synth $\to$ Real)}} \\[2pt]
\quad Finetune Synth $\to$ Real
  & 0.6798 ± 0.0208 & 0.4639 ± 0.0286 & 0.7991 ± 0.0963 & 0.5938 ± 0.0431
  & \cellcolor{gray!50}0.6903 ± 0.0261 & \cellcolor {yellow!40}0.4800 ± 0.0255 \\
\quad Finetune Synth (WB) $\to$ Real
  & \cellcolor {yellow!40}0.7227 ± 0.0233 & \cellcolor {yellow!40}0.5067 ± 0.0118 & 0.7642 ± 0.0260 & \cellcolor {yellow!40}0.6520 ± 0.0425
  & \cellcolor {yellow!40}0.6910 ± 0.0207 & \cellcolor{gray!50}0.4603 ± 0.0227 \\
\bottomrule
\end{tabular}%
}
\begin{minipage}{\linewidth}
\footnotesize
$^\dagger$\,Precision (P) and Recall (R) are Ultralytics-specific single-threshold metrics, not produced by LW-DETR's COCO API
\end{minipage}
\end{table}
\subsection{Experiments on the toy Ferrari datasets}
This section presents the quantitative results, qualitative observations, and ablation studies for the toy Ferrari datasets. All models are evaluated on a separate test set of real images that the models have not seen during training.
\noindent\textbf{Quantitative analysis}
Table~\ref{tab:combined_results} summarises detection performance across all
training regimes.
Training on the full real dataset (100\%) establishes the performance baseline,
with YOLO26 achieving mAP50\,=\,0.610 and mAP50-95\,=\,0.402, and LW-DETR
reaching mAP50\,=\,0.535 and mAP50-95\,=\,0.366, confirming a consistent
advantage for YOLO26 on this single-class dataset (see Table~\ref{tab:combined_results},
Baselines).
Synthetic-only training yields substantially lower scores for both models
(YOLO26: 0.451\,/\,0.300; LW-DETR: 0.382\,/\,0.243), reflecting the
\emph{domain gap} between rendered and real images.
Under scarce real-data conditions (10\% to 25\%), both models struggle to learn
meaningful detectors, with YOLO26 falling below mAP50\,=\,0.06 and LW-DETR
below 0.36.
Infusing synthetic images alongside real data provides a clear remedy, where at
10\% real data YOLO26 recovers to mAP50\,=\,0.507 (from 0.056 real-only) and
LW-DETR to 0.387 (from 0.238), with performance scaling consistently as the
real fraction increases.
At full infusion (100\%), both models surpass their real-only baselines
(YOLO26: 0.691\,/\,0.463; LW-DETR: 0.663\,/\,0.458), demonstrating that
synthetic data acts as a useful regulariser even when all real data is available
(see Table~\ref{tab:combined_results}, Synth + Real Mixed).
The best results are achieved by the two-stage fine-tuning strategy, where
\emph{Finetune Synth\,(WB) $\to$ Real} attains the best YOLO26 scores across
all metrics, except precision (mAP50\,=\,0.723, mAP50-95\,=\,0.507, P\,=\,0.764, R\,=\,0.652),
while for LW-DETR, \emph{Finetune Synth $\to$ Real} yields the highest
mAP50-95 (0.480) and the WB variant leads on mAP50 (0.691)
(see Table~\ref{tab:combined_results}, Fine-tuning).
\noindent\textbf{Ablation}
Table~\ref{tab:combined_results} includes paired white-background (WB)\,/\,non-WB runs and varying
real-data fractions, allowing two factors to be isolated.
\noindent\textit{Effect of random colored background:}
Since the real images were captured against a static white background,
synthetics rendered on non-white backgrounds carry an inherent appearance
mismatch.
Without any real data, synthetic data with white background still hurt performance
(YOLO26 mAP50: $-$0.045; LW-DETR: $-$0.056; see Table~\ref{tab:combined_results}, Baselines),
indicating that matching the background alone is insufficient without a
real-domain anchor.
Once real images are present the effect reverses: at 25\% infusion WB adds
$+$0.050 mAP50 for YOLO26 (0.598 vs.\ 0.548), with further gains at 50\% and
100\% ($+$0.038 and $+$0.015; see Table~\ref{tab:combined_results}, Synth + Real Mixed),
while LW-DETR remains largely unaffected.
The same pattern holds in the fine-tuning regime, where WB pre-training yields
$+$0.043 mAP50 for YOLO26 but negligible change for LW-DETR
($+$0.001 mAP50\,/\,$-$0.020 mAP50-95; see Table~\ref{tab:combined_results}, Fine-tuning),
suggesting the convolutional backbone is more sensitive to background appearance
than the transformer-based architecture.
\noindent\textit{Effect of real-data fraction in mixed training:}
Increasing the proportion of real images in the infused regime consistently
improves performance for both models (Table~\ref{tab:combined_results}, Synth + Real Mixed).
For YOLO26, mAP50 increases from 0.507 at 10\% real data to 0.548 at 25\%,
0.635 at 50\%, and 0.691 at 100\%. LW-DETR follows the same pattern, reaching
0.387, 0.512, 0.607, and 0.663, respectively. Across all real-data fractions,
mixed training clearly outperforms the corresponding real-only setting. In addition,
the 50\% infused configuration already surpasses the 100\% real-only baseline (mAP50) for
both YOLO26 (0.635 vs.\ 0.610) and LW-DETR (0.607 vs.\ 0.535), suggesting that
synthetic data offers a complementary benefit beyond simply compensating for
limited real-data availability.
\noindent\textbf{Qualitative analysis}
\begin{figure}[htbp]
    \centering
    \includegraphics[width=\textwidth]{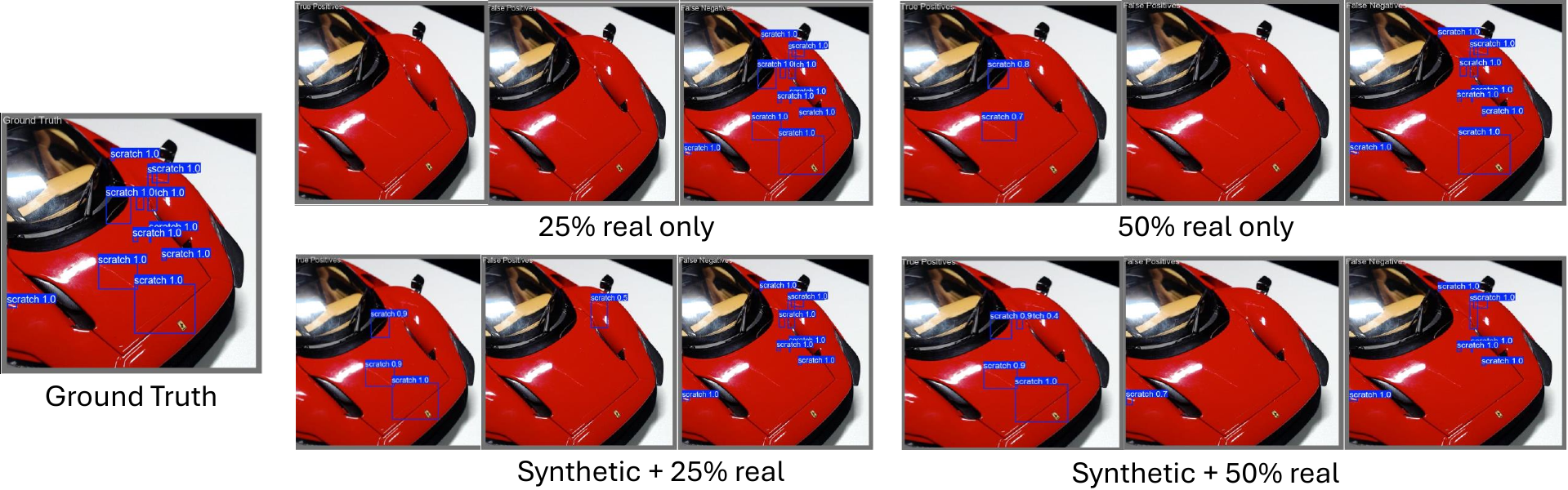}
    \caption{Qualitative detection results (toy Ferrari test set) comparing real-only versus synthetic-augmented training under scarce data conditions. Each column (left to right) shows a representative true positive, false positive, and false negative for four configurations: \textit{25\%
real only}, \textit{50\% real only}, \textit{Synthetic + 25\% real}, and
\textit{Synthetic + 50\% real}.}
    \label{fig:qualitative_toy_ferrari}
\end{figure}
Figure~\ref{fig:qualitative_toy_ferrari} compares real-only and synthetic-augmented training under scarce data conditions. With 25\% real data alone, the model fails to produce meaningful detections, resulting predominantly in false negatives. At 50\% real only, some detections appear but false negatives remain frequent. Infusing synthetic data at 25\% real already yields more true positive detections, and this further improves at 50\% infused, with false negatives decreasing in both cases. This indicates that combining synthetic and real data has a clear positive impact even when real training images are scarce.
\subsection{Experiments on the industrial grip datasets}
This section presents results for the industrial grip datasets using YOLOX-S~\cite{yolox2021},
selected for its permissive license suitable for commercial deployment.
All models are evaluated on a separate test set of real images that the models
have not seen during training.
\noindent\textbf{Training regime}
Four training regimes are evaluated: purely synthetic, mixed (synthetic + real),
fine-tuning (synthetic weights initialized and then trained on real data),
and real-only as a baseline. Four synthetic dataset variants are considered,
formed by combining two color modes (\texttt{tricolour}, \texttt{randomcolour})
with two camera modes (\texttt{randomcam}, \texttt{tripod}). Full details of each mode are provided in Section~\ref{sec:datasets}.
\begin{table}[htbp]
\centering
\caption{AP and AR on the industrial grip test set across all training regimes using YOLOX~\cite{yolox2021}. \colorbox{yellow!40}{Yellow} = best result,
\colorbox{gray!50}{Gray} = second best.}
\label{tab:grip_results}
\scalebox{0.7}{
\begin{tabular}{lcccccccc}
\toprule
& \multicolumn{2}{c}{\textbf{Real}} & \multicolumn{2}{c}{\textbf{Synthetic}} & \multicolumn{2}{c}{\textbf{Mixed}} & \multicolumn{2}{c}{\textbf{Fine-tuned}} \\
\cmidrule(lr){2-3}\cmidrule(lr){4-5}\cmidrule(lr){6-7}\cmidrule(lr){8-9}
\textbf{Training Regime} & \textbf{AP} & \textbf{AR} & \textbf{AP} & \textbf{AR} & \textbf{AP} & \textbf{AR} & \textbf{AP} & \textbf{AR} \\
\midrule
real baseline           & 0.234 & 0.317 & --                         & --                         & --                         & --                         & --                         & --                         \\
randomcolour\_randomcam & --    & --    & 0.089   & 0.145  & \cellcolor{gray!50}0.275  & \cellcolor{gray!50}0.359  & \cellcolor {yellow!40}0.319  & \cellcolor {yellow!40}0.389  \\
randomcolour\_tripod    & --    & --    & 0.011                       & 0.030                      & 0.217                      & 0.314                      & 0.282                      & 0.354                      \\
tricolour\_randomcam    & --    & --    & \cellcolor {yellow!40}0.196   & \cellcolor {yellow!40}0.243  & \cellcolor {yellow!40}0.316  & \cellcolor {yellow!40}0.396  & \cellcolor{gray!50}0.303  & \cellcolor{gray!50}0.384  \\
tricolour\_tripod       & --    & --    & \cellcolor{gray!50}0.110   & \cellcolor{gray!50}0.159  & 0.237  & 0.330  & 0.275                      & 0.359                      \\
\bottomrule
\end{tabular}
}
\end{table}
%

%
\noindent\textbf{Quantitative analysis}
Table~\ref{tab:grip_results} summarizes average precision (AP) and average recall (AR) on the industrial grip test
set of real images for all four regimes. For the mixed training
results reported here, all available real images (120) were used; the effect of
incrementally adding real data is examined separately in the ablation study. Training on synthetic
data alone falls below the real baseline in all configurations.
\texttt{tricolour\_randomcam} comes closest with AP~0.196, while
\texttt{randomcolour\_tripod} collapses to AP~0.011, highlighting the importance
of randomised camera viewpoints as well as realistic coloring. Mixed training recovers performance across all
variants and surpasses the real baseline clearly when \texttt{randomcam} variants are
used, with \texttt{tricolour\_randomcam} reaching AP~0.316 and AR~0.396.
Fine-tuning proves to be the most consistent strategy, with all four variants
exceeding the real baseline. \texttt{randomcolour\_randomcam} achieves the best
overall result of AP~0.319 and AR~0.389, and even the weaker \texttt{tripod} variants
yield a meaningful gain over training on real
data alone.
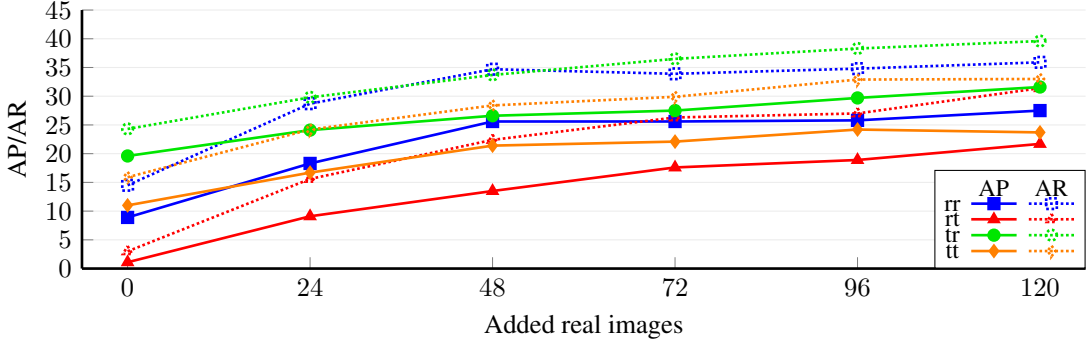
\begin{figure}[htbp]
    \centering
    \begin{tikzpicture}
        \begin{axis}[
            name=plot,
            width=0.9\linewidth,
            height=5cm,
            xlabel={Added real images},
            ylabel={AP/AR},
            xmin=-6, xmax=126,
            ymin=0, ymax=45,
            xtick={0, 24, 48, 72, 96, 120},
            ytick={0, 5, 10, 15, 20, 25, 30, 35, 40, 45},
            axis y line*=left,
            axis x line*=bottom,
            xmajorgrids=false,
            xminorgrids=false,
            ymajorgrids=true,
            yminorgrids=false,
            grid style={solid, gray!15},
            mark size=2pt,
            line width=1pt,
            legend image post style={mark size=2pt, line width=1pt},
        ]

        \addplot[color=blue, mark=square*, solid] coordinates {
            (0,8.9)(24,18.3)(48,25.6)(72,25.6)(96,25.8)(120,27.5)};
        \label{plot:ap_rr}

        \addplot[color=blue, mark=square, densely dotted] coordinates {
            (0,14.5)(24,28.7)(48,34.7)(72,33.9)(96,34.8)(120,35.9)};
        \label{plot:ar_rr}

        \addplot[color=red, mark=triangle*, solid] coordinates {
            (0,1.1)(24,9.1)(48,13.5)(72,17.6)(96,18.9)(120,21.7)};
        \label{plot:ap_rt}

        \addplot[color=red, mark=triangle, densely dotted] coordinates {
            (0,3)(24,15.6)(48,22.4)(72,26.3)(96,27)(120,31.4)};
        \label{plot:ar_rt}

        \addplot[color=green!90!black, mark=*, solid] coordinates {
            (0,19.6)(24,24.1)(48,26.6)(72,27.5)(96,29.7)(120,31.6)};
        \label{plot:ap_tr}

        \addplot[color=green!90!black, mark=o, densely dotted] coordinates {
            (0,24.3)(24,29.8)(48,33.7)(72,36.5)(96,38.3)(120,39.6)};
        \label{plot:ar_tr}

        \addplot[color=orange, mark=diamond*, solid] coordinates {
            (0,11)(24,16.7)(48,21.4)(72,22.1)(96,24.2)(120,23.7)};
        \label{plot:ap_tt}

        \addplot[color=orange, mark=diamond, densely dotted] coordinates {
            (0,15.9)(24,24.2)(48,28.4)(72,29.9)(96,32.9)(120,33)};
        \label{plot:ar_tt}

        \end{axis}

        \matrix[
            draw,
            matrix of nodes,
            anchor=south east,
            font=\small,
            fill=white,
            inner sep=1.8pt,
            row sep=-2.8pt,
        ] at (plot.south east) {
            & AP & AR \\
            rr & \ref{plot:ap_rr} & \ref{plot:ar_rr} \\
            rt & \ref{plot:ap_rt} & \ref{plot:ar_rt} \\
            tr & \ref{plot:ap_tr} & \ref{plot:ar_tr} \\
            tt & \ref{plot:ap_tt} & \ref{plot:ar_tt} \\
        };
    \end{tikzpicture}
\caption{AP and AR as real images are incrementally added during mixed training
for each training regime (rr: \texttt{randomcolour\_randomcam},
rt: \texttt{randomcolour\_tripod}, tr: \texttt{tricolour\_randomcam},
tt: \texttt{tricolour\_tripod}). The AP values are presented in this plot as percentages for better visibility.}
\label{fig:infused_training}
\end{figure}
\noindent\textbf{Ablation} The effect of incrementally adding real images to the synthetic training set is
examined in Figure~\ref{fig:infused_training}. A consistent trend is observed
across all variants, where performance increases with more real data, though
marginal gains diminish in later stages. \texttt{randomcolour\_randomcam}
(\textbf{rr}), while performing poorly as a purely synthetic model, improves
substantially during the first two stages (up to 48 images), reaching
performance comparable to \texttt{tricolour\_randomcam} (\textbf{tr}) by that
point. \textbf{tr}, which starts with superior metrics, improves more gradually
but continues to gain throughout all stages, reaching only around 60\% of its
total improvement by stage two, compared to 90--95\% for \textbf{rr}. We
hypothesise that this difference stems from dataset balance. Since \textbf{tr}
closely resembles real images, each additional sample further closes the domain
gap without disturbing the distribution. \textbf{rr}, by contrast, benefits
greatly from early real examples but begins to stagnate once real images exceed
roughly 48, at which point they start to overrepresent the real domain and
destabilize the mixed dataset.
\begin{figure}
    \centering
    \includegraphics[width=0.86\textwidth]{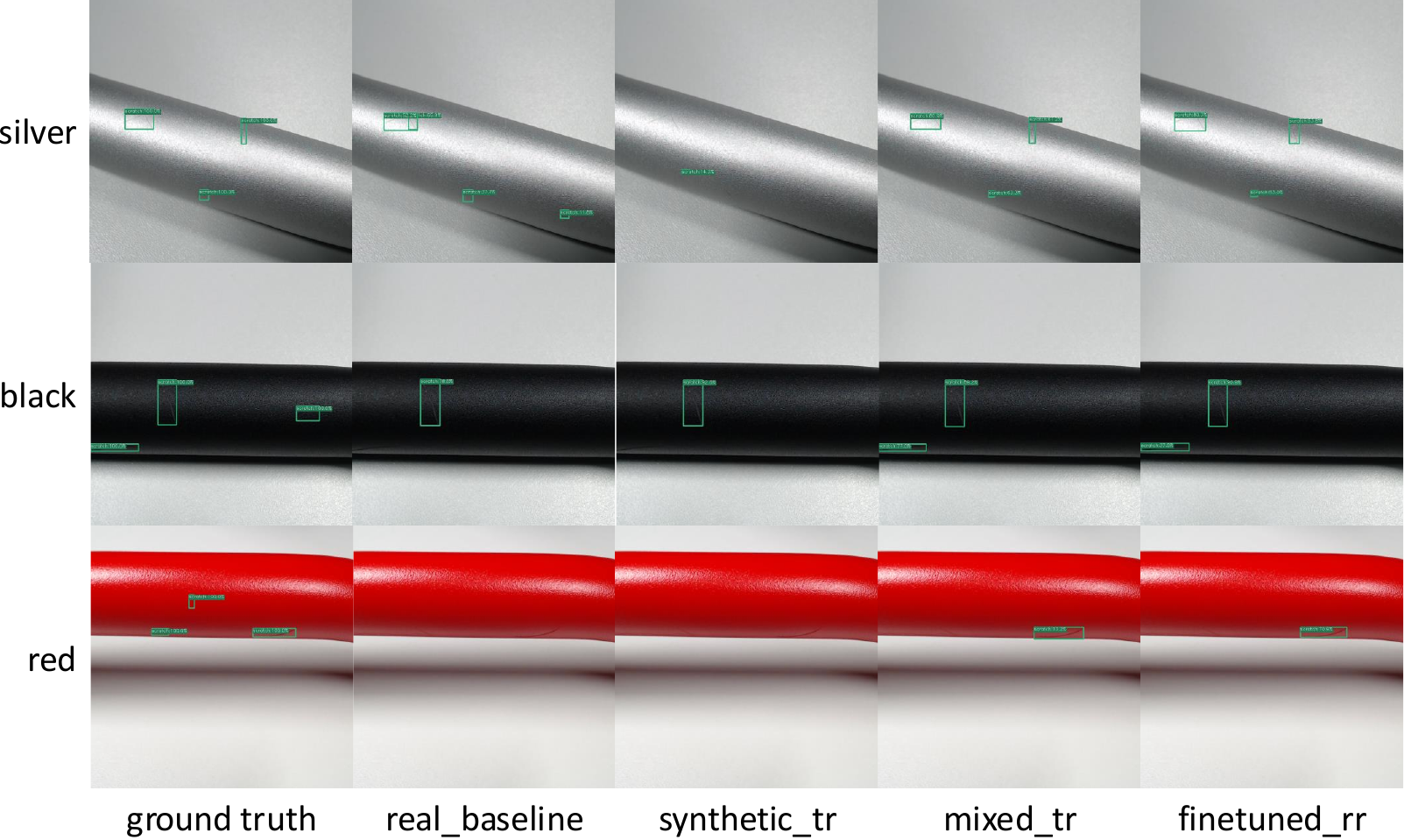}
    \caption{Qualitative results using industrial grip datasets. All images were created using a confidence threshold of 0.1 and a NMS (non-maximum suppression) threshold of 0.3.}
    \label{fig:compared_images}
\end{figure}
\noindent\textbf{Qualitative analysis} In addition to the quantitative
analysis, Figure~\ref{fig:compared_images} shows a selection of qualitative
results across all three objects. For the silver object, the two combined
training strategies closely mirror the ground-truth annotations, while \texttt{real\_baseline} produces a false positive and a false negative, and \texttt{synthetic\_tr} only produces a false positive. For the black object, one scratch is found by
all models, a second is missed only by \texttt{real\_baseline} and \texttt{synthetic\_tr}, and the third is missed
entirely. The red object
represents the most difficult case, where two scratches go undetected across all
models and \texttt{real\_baseline} and \texttt{synthetic\_tr} missing the third as well. Overall, the
results highlight clear room for improvement while demonstrating the positive
impact of combining real and synthetic data during training.
\section{Conclusion}
This paper presented a procedural rendering pipeline for annotated synthetic
scratch data generation and evaluated four training strategies across two
objects with differing material properties and three lightweight detectors.
Results consistently show that synthetic-only training falls short due to the
domain gap, yet synthetic data remains highly beneficial when combined with
real images. Mixed training effectively recovers performance under scarce
real-data conditions, while fine-tuning from synthetic weights proved the most
robust strategy, outperforming real-only training across all architectures and
datasets.

Despite these gains, fine and low-contrast scratches remain challenging across
all strategies. Future work could explore higher-fidelity material simulation,
advanced domain adaptation techniques, and active learning to prioritize the
most informative real samples. The procedural nature of the pipeline also opens
the door to simulating defects under challenging real-world conditions such as
dust, mud, and contaminated or weathered surfaces, which are difficult and
costly to capture in real annotated data. Extending the pipeline to other defect
types and object geometries would further validate its generality.

\noindent\textbf{Acknowledgement} This project was partially funded by the EU Horizon Europe project dAIEDGE under grant agreement No. 101120726.
\appendix
\section*{\centering Appendix}
\section{Automatic annotations using our procedural pipeline}
\label{app:annotations}

One of the major advantages of synthetic data is the ability to generate 
annotations automatically in the data generation pipeline. More precisely, it 
is necessary to know which pixels depict scratches and to use this information 
to create annotation data.

The applied scratch mask contains exactly this information, as it directly 
encodes which pixels of the resulting image contain scratches. The 
\emph{AOV Output} node in the material graph is responsible for supplying this 
information after rendering. AOV stands for ``Arbitrary Output Variables''; the 
node receives the applied scratch map as input and exposes it as an output 
accessible after rendering.

\begin{figure}[htbp]
    \centering
    \includegraphics[width=0.25\linewidth]{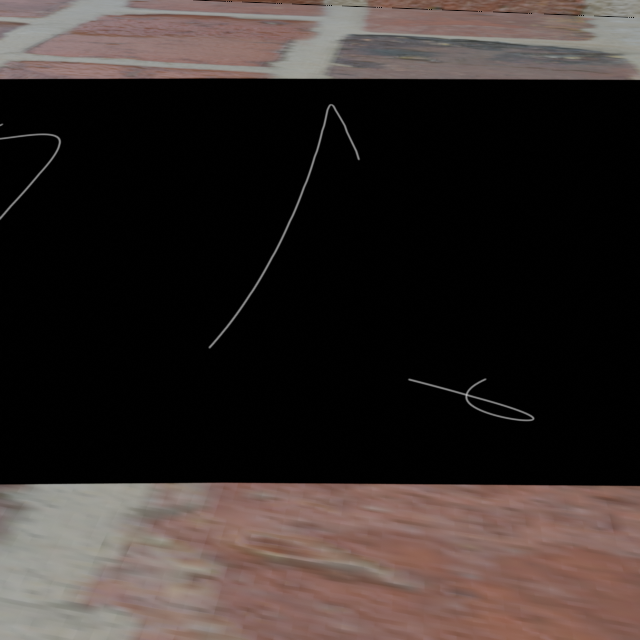}%
    \includegraphics[width=0.25\linewidth]{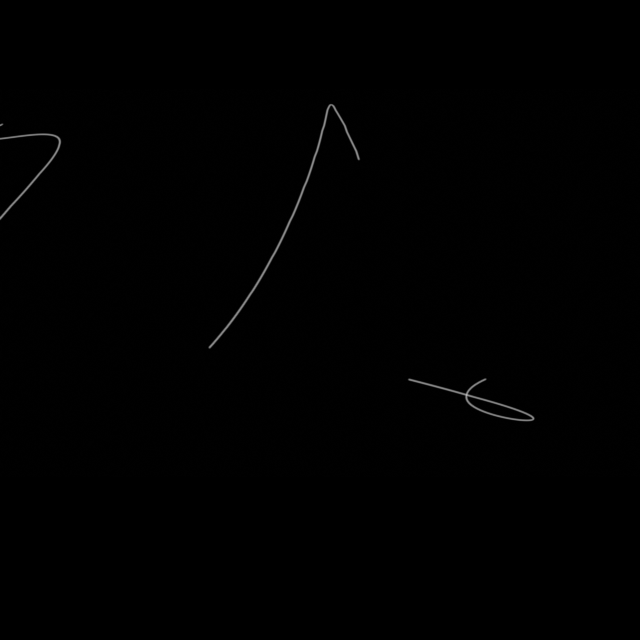}%
    \includegraphics[width=0.25\linewidth]{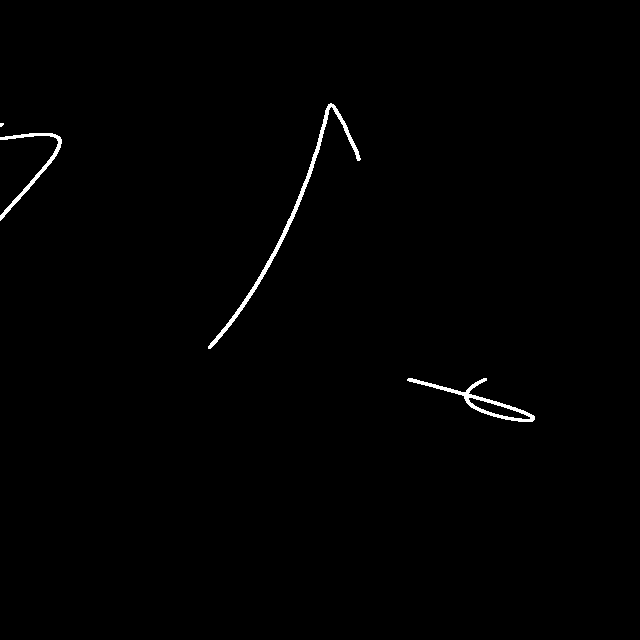}%
    \includegraphics[width=0.25\linewidth]{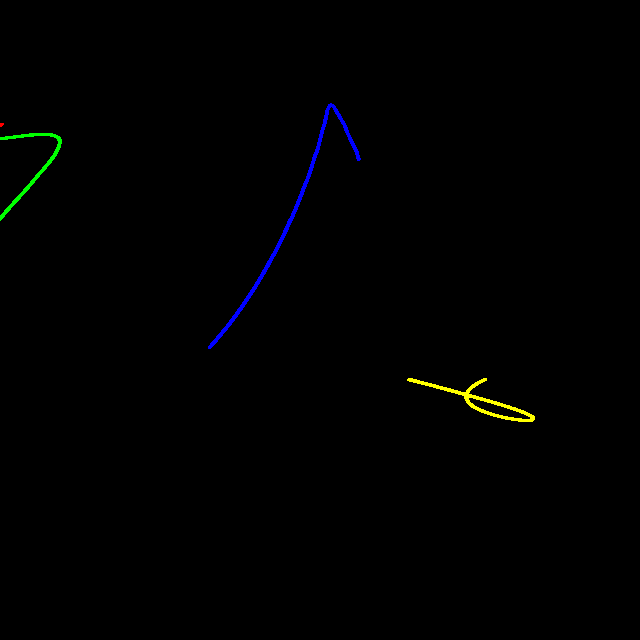}
    \caption{Stages of scratch annotation (left to right): scratch positions 
             in the scene, as provided by the AOV Output node, converted to a 
             binary representation, and individually labeled scratches.}
    \label{fig:AOV_scratches}
\end{figure}

The image data provided by the AOV node only returns the applied mask, 
coloring everything else black, as shown in \cref{fig:AOV_scratches}. To 
extract the needed information, the RGB image is converted into an array 
encoding each scratch pixel as $1$ and everything else as $0$. In practice, an 
array of the same size as the image is initialized with ones, and every pixel 
with a value of $(0,0,0)$ or $(1,1,1)$ is set to zero. A labeling function 
then assigns a unique label to each connected island of ones, so that each 
pixel is identified as either background or a specific scratch instance. This 
segmentation data is passed to the COCO annotation writer implemented in 
BlenderProc~\cite{denninger2023blenderproc2}, which produces a \texttt{.json} file containing all annotations, 
including bounding boxes, required for detection tasks.
\begin{figure}[htbp]
    \centering
    \begin{subfigure}{0.48\linewidth}
        \includegraphics[width=\textwidth]{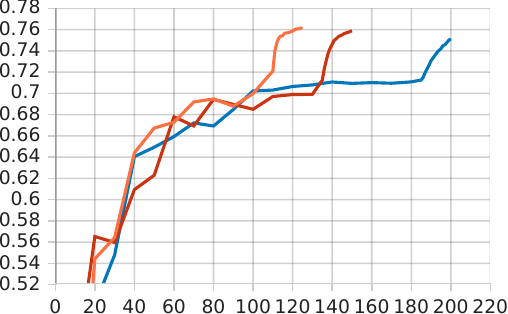}
        \caption*{AP}
    \end{subfigure}
    \begin{subfigure}{0.48\linewidth}
        \includegraphics[width=\textwidth]{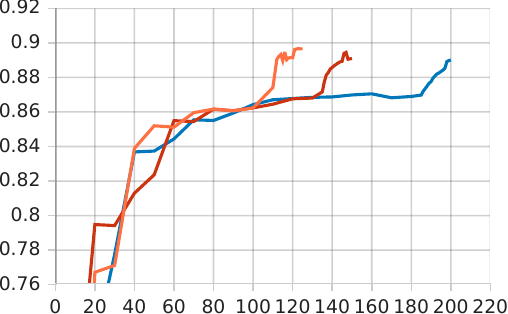}
        \caption*{$\text{AP}_{50}$}
    \end{subfigure}
    \caption{AP (left) and $\text{AP}_{50}$ (right) of \texttt{synthetic\_rr} 
             on the industrial grip dataset, showing stagnation around epoch~125.}
    \label{fig:ap_stagnation_graphs}
\end{figure}
\section{Training Epoch Selection}
\label{app:stagnation}
To avoid overfitting to the synthetic domain, training runs were monitored by 
tracking AP and $\text{AP}_{50}$ on the validation set throughout training. 
Figure~\ref{fig:ap_stagnation_graphs} shows the stagnation curves for the 
\texttt{synthetic\_rr} experiment on the industrial grip dataset. Both metrics 
plateau around epoch~125, after which gains become negligible. This value was 
therefore chosen as the maximum number of training epochs for all experiments. 
Analogous behaviour was observed across all other configurations.
\begin{figure}[htbp]
    \centering
    \includegraphics[width=\linewidth]{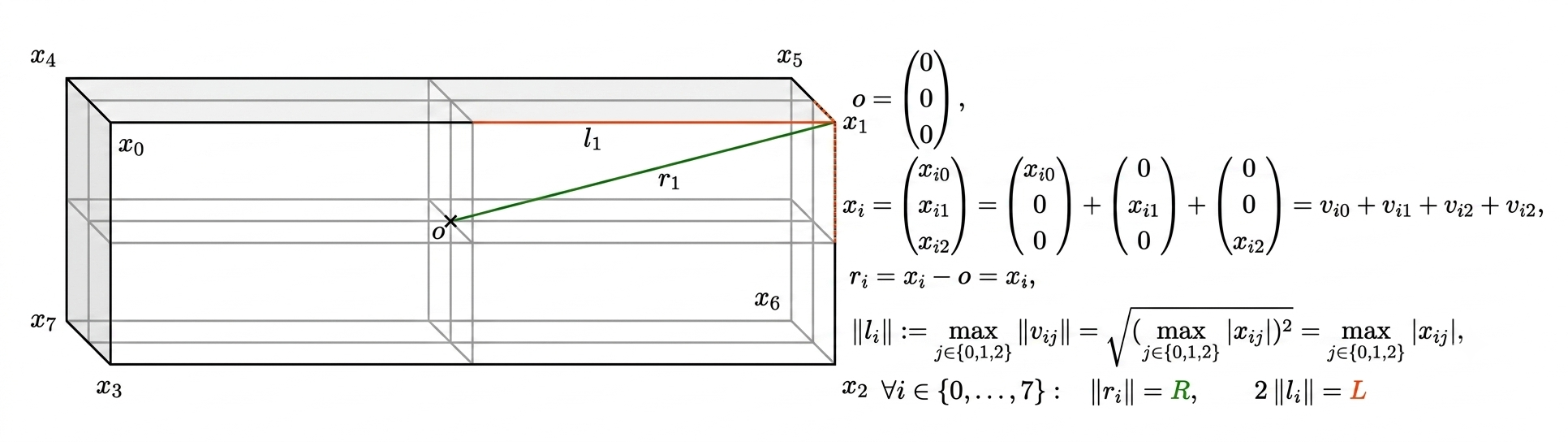}
    \caption{Reach and length computed from the bounding box after setting 
             the origin to its centre.}
    \label{fig:reach_length}
\end{figure}
\section{Object Positioning via Bounding Box}
\label{app:bounding_box}
To correctly place and scale the object in the scene, its origin is set to the 
centre of its bounding box. All corner coordinates are then expressed relative 
to this centre in object space, making the geometry symmetric about the origin.

The model's length is computed as twice the maximum absolute corner coordinate. 
Dividing the real-world object length, supplied by the user via \\
\texttt{object\_length}, by this value gives the scale factor required for a 
1:1 ratio. The object's \emph{reach} is defined as the maximum distance it can 
extend from its origin in any direction, equal to the distance from the origin 
to any bounding box corner multiplied by the scale factor to convert to world 
space. This quantity is used throughout the pipeline to prevent the object from 
clipping into surrounding geometry. Both values are illustrated in 
\cref{fig:reach_length}.
\section{Qualitative results}
We present additional qualitative results on the toy Ferrari dataset to further illustrate how combining synthetic and real data, or fine-tuning on synthetic weights before real training, improves detection performance compared to training on real data alone, using YOLO26-Nano and LW-DETR-Tiny.
\begin{figure}[htbp]
    \centering
    \includegraphics[width=\textwidth]{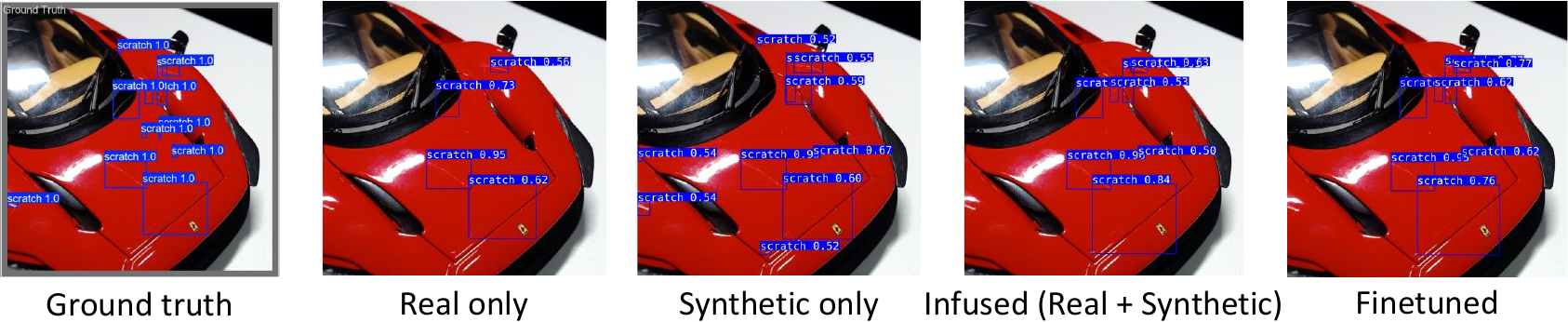}
    \caption{Qualitative detection results of LW-DETR-Tiny~\cite{chen2024lw} on the toy Ferrari 
dataset with a confidence threshold of 0.5. Columns show, from left to right: 
ground truth, real only (100\%), synthetic only, infused (Synthetic~+~Real 
100\%, white background), and fine-tuned (Synthetic\,(WB)~$\to$~Real). 
}
\label{fig:qualitative_lwdetr}
\end{figure}

 Figure~\ref{fig:qualitative_lwdetr} shows qualitative predictions from LW-DETR-Tiny across four training configurations. The real-only baseline produces the weakest results, with frequent missed detections. Synthetic-only training performs noticeably better, yielding more true positives. Infused training (Synthetic + Real 100\%, WB) produces similar results to synthetic only, while fine-tuning achieves comparable performance with a slight reduction in false negatives, making it the most reliable configuration overall based on these results.

\begin{figure}[htbp]
    \centering
    \includegraphics[width=\textwidth]{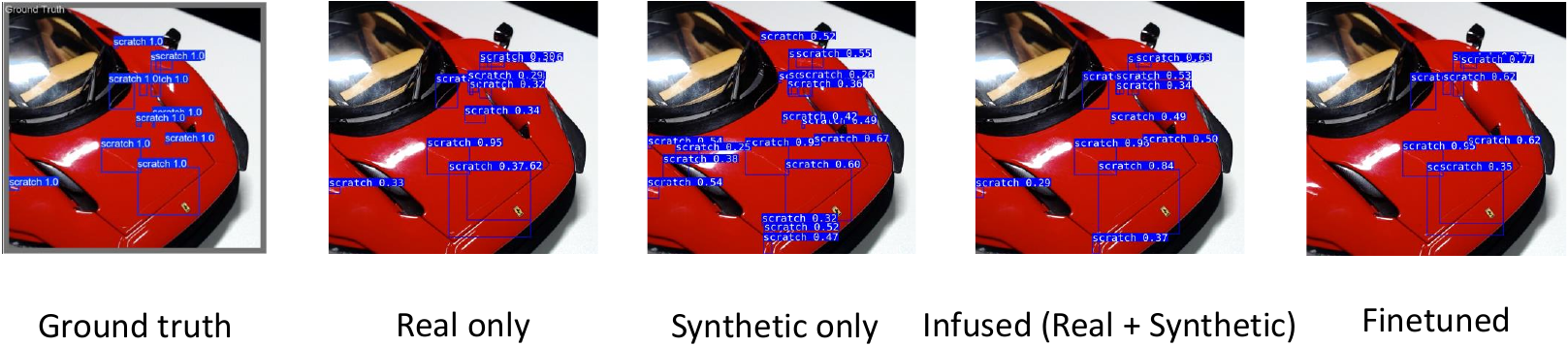}
    \caption{Qualitative detection results of LW-DETR-Tiny~\cite{chen2024lw} on the toy Ferrari 
dataset with a confidence threshold of 0.25. Columns show, from left to right: 
ground truth, real only (100\%), synthetic only, infused (Synthetic~+~Real 
100\%, white background), and fine-tuned (Synthetic\,(WB)~$\to$~Real). 
}
    \label{fig:qualitative_lwdetr_lowerconfidence}
\end{figure}

\begin{figure}[htbp]
    \centering
    \includegraphics[width=\textwidth]{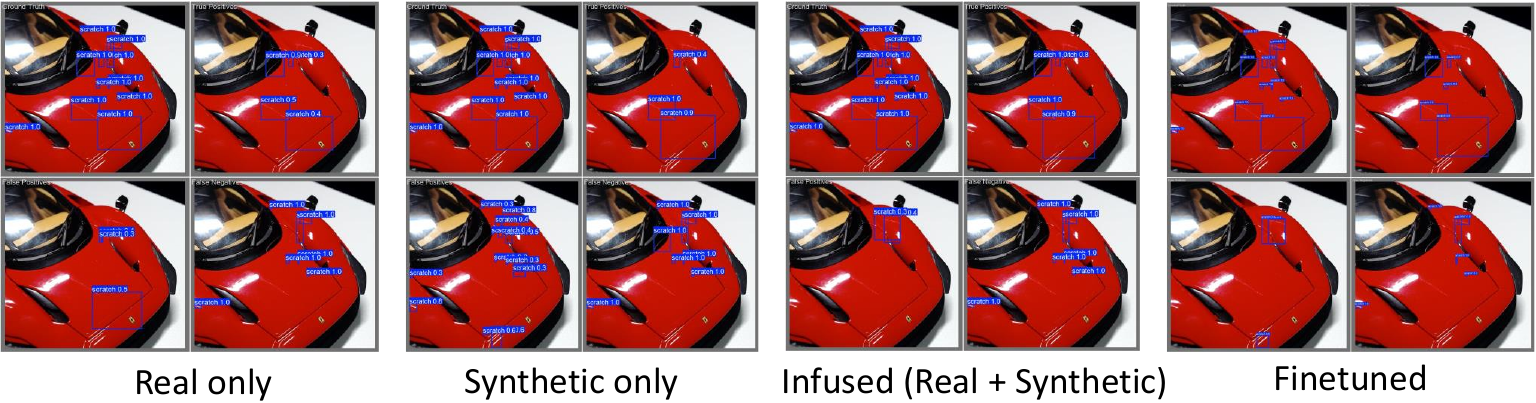}
    \caption{Qualitative detection results of YOLO26-nano~\cite{jocher2026ultralyticsyolo26unifiedrealtime} on the toy Ferrari 
dataset with a confidence threshold of 0.25. Columns show, from left to right: 
real only (100\%), synthetic only, infused (Synthetic~+~Real 
100\%, white background), and fine-tuned (Synthetic\,(WB)~$\to$~Real). 
}
\label{fig:qualitative_yolo26}
\end{figure}

Figure~\ref{fig:qualitative_lwdetr_lowerconfidence} shows the same configurations at a reduced confidence threshold of 0.25. While lowering the threshold recovers some missed detections, it also introduces a notable increase in false positives across all configurations, indicating that LW-DETR-Tiny produces less calibrated confidence scores compared to YOLO26-Nano (see Figure~\ref{fig:qualitative_yolo26}).

\begin{figure}[htbp]
    \centering
    \includegraphics[width=\textwidth]{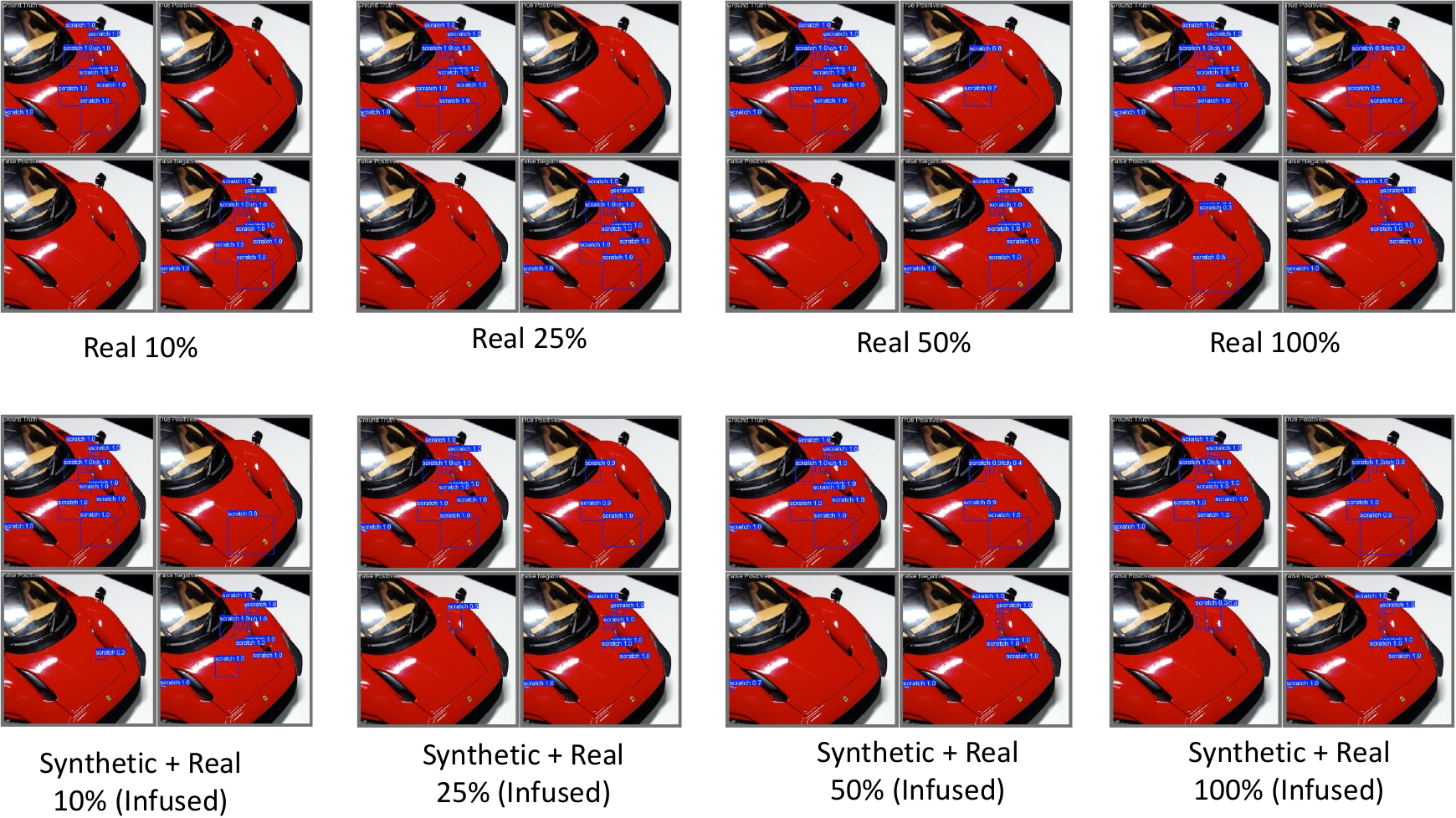}
    \caption{Qualitative detection results of YOLO26-Nano~\cite{jocher2026ultralyticsyolo26unifiedrealtime} on the toy Ferrari 
dataset. Columns show, from left to right: Real 10\%, Real 25\%, Real 50\%, 
Real 100\%, Synthetic~+~Real 10\% (Infused), Synthetic~+~Real 25\% (Infused), 
Synthetic~+~Real 50\% (Infused), and Synthetic~+~Real 100\% (Infused). 
}
\label{fig:real_scarce_full}
\end{figure}
Figure~\ref{fig:qualitative_yolo26} shows qualitative predictions from YOLO26-Nano across the same four configurations. The results follow the same trend as LW-DETR-Tiny, with synthetic-only outperforming real-only and infused and fine-tuned configurations producing comparable results. However, YOLO26-Nano exhibits fewer false negatives overall, reflecting its stronger detection calibration at this threshold.

We present the full qualitative results of the scarce data analysis in Figure~\ref{fig:real_scarce_full}, covering all real-data fractions (10\%, 25\%, 50\%, and 100\%) alongside their synthetic-infused counterparts.

\bibliographystyle{unsrt}  
\bibliography{references}  

\end{document}